\documentclass[runningheads]{llncs}
\usepackage{url}
\usepackage{graphicx}
\usepackage{amsmath,amssymb}
\usepackage{float}
\usepackage{comment}
\usepackage[colorinlistoftodos]{todonotes}
\newcommand*\samethanks[1][\value{footnote}]{\footnotemark[#1]}
\DeclareMathOperator{\ex}{\mathbb{E}}
\newcommand{\ptheta}{p_{\boldsymbol{\theta}}}
\newcommand{\qphi}{q_{\boldsymbol{\phi}}}
\newcommand{\x}{\boldsymbol{x}}
\newcommand{\z}{\boldsymbol{z}}
\newcommand{\thetab}{\boldsymbol{\theta}}
\newcommand{\phib}{\boldsymbol{\phi}}

\begin{document}
\title{Semi-Supervised Variational Autoencoder for Survival Prediction}
%
%
\author{Sveinn P\'{a}lsson\thanks{Authors contributed equally.}\inst{1} \and
Stefano Cerri\samethanks\inst{1} \and
Andrea Dittadi\samethanks\inst{2} \and
Koen Van Leemput\inst{1,3}}
%
%
\institute{Department of Health Technology, Technical University of Denmark, Denmark
\and
Department of Applied Mathematics and Computer Science, Technical University of Denmark, Denmark 
\and
Athinoula A. Martinos Center for Biomedical Imaging, Massachusetts General Hospital, Harvard Medical School, USA}
\maketitle              
\begin{abstract}
In this paper we propose a semi-supervised variational autoencoder for classification of overall survival groups from tumor segmentation masks. The model can use the output of any tumor segmentation algorithm, removing all assumptions on the scanning platform and the specific type of pulse sequences used, thereby increasing its generalization properties. Due to its semi-supervised nature, the method can learn to classify survival time by using a relatively small number of labeled subjects. We validate our model on the publicly available dataset from the Multimodal Brain Tumor Segmentation Challenge (BraTS) 2019.

\end{abstract}

\section{Introduction}

Brain tumor prognosis involves forecasting the future disease progression in a patient, which is of high potential value for planning the most appropriate treatment. Glioma is the most common primary brain tumor and patients suffering from its most aggressive form, glioblastoma, have generally very poor prognosis. Glioblastoma patients have a median overall survival (OS) of less than 15 months, and a 5-year OS rate of only 10\% even when they receive treatment \cite{fetpet}.
Automatic prediction of overall survival of glioblastoma patients is an important but unsolved problem, with no established method available in clinical practice. 

The last few years have seen an increased interest in brain tumor survival time prediction from magnetic resonance (MR) images, often using discriminative methods that directly encode the relationship between image intensities and prediction labels~\cite{BRATS2018}. 
However, due to the flexibility of MR imaging, such methods do not generalize well to images acquired at different centers and with different scanners, limiting their potential applicability in clinical settings. 
Furthermore, being supervised methods, they require ``labeled'' training data where for each training subject both imaging data and ultimate survival time are available. 
Although public imaging databases with survival information have started to be collected~\cite{BRATSOriginal,BRATS2,bakas_citation,bakas_citation_2}, 
the requirement of such labeled data fundamentally limits the number of subjects available for training, severely restricting the prediction performance attainable with current methods.

In this paper, we explore whether the aforementioned issues with supervised intensity-based methods can be ameliorated by using a semi-supervised approach instead, using only segmentation masks as input. In particular, we adapt a semi-supervised variational autoencoder model~\cite{VAESEMIKingma} to predict overall survival from a small amount of labeled training subjects, augmented with \emph{unlabeled} subjects in which only imaging data is available. 
The method only takes segmentation masks as input, thereby removing all assumptions on the image modalities and scanners used.

The Multimodal Brain Tumor Segmentation Challenge (BraTS) \cite{BRATSOriginal} has been held every year since 2012, and focuses on the task of segmenting three different brain tumors structures (``enhancing tumor'', ``tumor core'' and ``whole tumor'') and ``background'' from multimodal MR images. Since 2017, BraTS has also included the task of OS prediction. In this paper we focus on the latter, classifying the scans into three prognosis groups:  \textbf{long-survivors} (\textgreater 15 months), \textbf{short-survivors} (\textless 10 months), and \textbf{mid-survivors} (between 10 and 15 months), all relative to the time of diagnosis. 


\section{Model}
\label{sec:Models}

We begin by formally describing the problem we aim to solve. 
The available training data consists of a set of $N_l$ labeled pairs 
$\{(\x_1,y_1),...,(\x_{N_l}, y_{N_l}) \} $,
possibly augmented with a set of $N_u$ \emph{unlabeled} data points 
$\{\x_{N_l+1},...,\x_{N_l+N_u} \}$,
where $\x_i \in \{1,...,M_x \}^D$ is the i-th subject's image data in the form of a segmentation map with $D$ voxels, and the target variable $y_i \in \{1,...,M_y \}$ denotes the survival group the subject belongs to. In our case we have 
the segmentation of
$M_x = 4$ different tumor structures as input to the model, and $M_y = 3$ different survival groups. For convenience, we will omit the index $i$ when possible in the remainder.

\begin{figure}
    \centering
    \includegraphics[width=.35\textwidth]{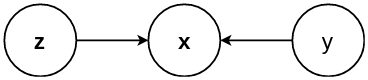}
    \caption{Probabilistic graphical model of the generative process.}
    \label{fig:semiVAE}
\end{figure}

We assume that the data is generated by a random process, illustrated in Figure~\ref{fig:semiVAE}, that involves some latent variables $\z \in \mathcal{R}^L$, assumed to be independent of $y$, where $L \ll D$. These latent variables encode high-level tumor shape and location features shared across survival groups.
Specifically, we assume a generative model of the form
\begin{equation}
  \ptheta(\x, y, \z) = \ptheta(\x|y,\z)p(\z)p(y),
\end{equation}
where $p({\z}) = \mathcal{N}({\z} | {\bf 0}, {\bf I})$ is a zero-mean isotropic multivariate Gaussian, $p(y) \propto 1$ is a flat categorical prior distribution over $y$, and 
$\ptheta(\x|y,\z)$ is a conditional distribution parameterized by $\thetab$. 

Our task is to find the maximum likelihood parameters, i.e., the parameter values $\thetab$ that maximize the probability of the training data under the model. This is equivalent to maximizing 
\begin{equation}
  \sum_{i=1}^{N_l} \log \ptheta( \x_i, y_i )
  +
  \sum_{i=N_l+1}^{N_l+N_u} \log \ptheta( \x_i )
  \label{eq:objectiveFunction}
\end{equation}
with respect to $\thetab$,
where
\begin{equation}
    \ptheta(\x,y) = \int_{\z} \ptheta(\x, y, \z)d\z
    \label{eq:marginal}
\end{equation}
and 
\begin{equation}
  \ptheta(\x) = \sum_y \ptheta(\x,y)
  .
\end{equation}
Once suitable parameter values are found, the survival group of a new subject with image data $\x$ can be predicted by 
assessing
$\ptheta(y | \x ) = \ptheta(\x,y) / \ptheta(\x)$.


\subsection{Semi-supervised variational autoencoder}


Maximizing eq.~\eqref{eq:objectiveFunction} 
for $\thetab$
directly is not feasible due to intractability of the integral over the latent variables 
in eq.~\eqref{eq:marginal}. 
We therefore use an
Expectation-Maximization (EM) \cite{EM} algorithm to
exploit the fact that 
the
optimization would be easier if 
the latent variables
were known.
The algorithm iteratively constructs and maximizes a
lower bound to eq.~\eqref{eq:objectiveFunction}
in a process that involves 
``filling in'' the 
missing
latent variables using their posterior distribution.
Since this posterior distribution is intractable,
we follow~\cite{VAESEMIKingma} and 
approximate 
$\ptheta( \z,y|\x)$ using a specific functional form 
$\qphi(\z|\x,y)$ with parameters $\phib$:
\begin{equation*}
    \qphi(\z,y|\x) = \qphi(\z|\x,y)\qphi(y|\x)
    ,
\end{equation*}
where $\qphi(\z|\x,y)$ is a multivariate Gaussian distribution with diagonal covariance matrix, and $\qphi(y|\x)$ is a categorical distribution.
%
%
This approximation can be used to obtain a lower bound to 
eq.~\eqref{eq:objectiveFunction}
as follows.
The probability of each \emph{labeled} data point (first term in eq.~\eqref{eq:objectiveFunction}) can be rewritten as:
\begin{align*}
    \log \ptheta(\x,y) &= \ex_{\qphi(\z|\x,y)}[\log \ptheta(\x,y)] \nonumber \\
     &= \ex_{\qphi(\z|\x,y)}\Big [\log \Big [ \frac{\ptheta(\x,y,\z)}{\ptheta(\z|\x,y)}\Big ]\Big ] \nonumber \\
     &= \ex_{\qphi(\z|\x,y)}\Big [\log \Big [ \frac{\ptheta(\x,y,\z)}{\qphi(\z|\x,y)}\frac{\qphi(\z|\x,y)}{\ptheta(\z|\x,y)}\Big ]\Big ] \nonumber \\
     &= \underbrace{\ex_{\qphi(\z|\x,y)}\Big [\log \Big [ \frac{\ptheta(\x,y,\z)}{\qphi(\z|\x,y)}\Big ]\Big ]}_{=\mathcal{L}_{\thetab, \phib}(\x,y)}+ \underbrace{\ex_{\qphi(\z|\x,y)}\Big [\log \Big [ \frac{\qphi(\z|\x,y)}{\ptheta(\z|\x,y)}\Big ]\Big ]}_{=D_{KL}(\qphi(\z|\x,y)||\ptheta(\z|\x,y))} 
\end{align*}
where $D_{KL}$ denotes the Kullback-Leibler (KL) divergence. Since the KL divergence is always non-negative, we have that
\begin{equation}
    \log \ptheta(\x,y) \geq \mathcal{L}_{\thetab, \phib}(\x,y)
    \label{eq:labeled_obj}
    .
\end{equation}
%
Using a similar derivation, the 
probability 
of each \emph{unlabeled} data point can be bounded as follows:
\begin{align}
    \log \ptheta(\x) &\geq \ex_{\qphi(y,\z|\x)}\Big [ \log \frac{\ptheta(\x,y,\z)}{\qphi(\z|y,\x)} - \log \qphi(y|\x)\Big ]\nonumber \\
    &= \sum_y \qphi(y|\x)(\mathcal{L}_{\thetab, \phib}(\x,y))+\mathcal{H}(\qphi(y|\x)) = \mathcal{U}_{\thetab,\phib}(\x),
    \label{eq:unlabeled_obj}
\end{align}
where $\mathcal{H}(\cdot)$ denotes the entropy of a probability distribution.

By combining \eqref{eq:labeled_obj} and \eqref{eq:unlabeled_obj},
a lower bound to 
eq.~\eqref{eq:objectiveFunction} is finally obtained as:
\begin{align}
    \mathcal{J}_{\thetab, \phib} = \sum_{i=1}^{N_l} \mathcal{L}_{\thetab, \phib}(\x_i,y_i) + 
    \sum_{i=N_l+1}^{N_l+N_u} \mathcal{U}_{\thetab, \phib}({\x_i})
     ,
    \label{eq:comb_objective}
\end{align}
which we optimize with respect to both the variational parameters $\phib$ and the generative parameters $\thetab$.
We use stochastic gradient ascent for the optimization, approximating gradients of the expectations in~\eqref{eq:comb_objective} as described in~\cite{VAEKingma}. Implementation details are discussed in Section~\ref{sec:implementation}.

From a information theory point of view, the latent unobserved variables $\z$ can be interpreted as a code. Therefore, we can refer to the 
distributions
$\qphi(\z|\x,y)$ and 
$\ptheta(\x | y, \z)$ as a probabilistic \emph{encoder} and \emph{decoder}, respectively~\cite{VAEKingma}. The label predictive distribution $\qphi(y|\x)$ has the form of a discriminative \emph{classifier}, and can be used 
as an approximation to $\ptheta(y|\x)$ for classifying new cases after training.

\subsection{Model modifications}

Here we describe a few model modifications for making the parameter learning process faster and less prone to overfitting. 

\subsubsection{Classification objective}
%
Note that in the objective function \eqref{eq:comb_objective}, the label predictive approximation $\qphi(y  |{\x})$ only appears in the bound for unlabeled data. To let $\qphi(y  |{\x})$ also learn from labeled data, we follow \cite{VAESEMIKingma} and add a weak classification loss, resulting in the modified objective
\begin{align}
    \mathcal{J}^{\alpha}_{\thetab, \phib} = \mathcal{J}_{\thetab, \phib} + \alpha  \sum_{i=1}^{N_l} \log q_{\phi}({y_i}|{\x_i})
\end{align}
where $\alpha$ controls the relative weight between generative and purely discriminative learning.

\subsubsection{Gumbel-Softmax}

One of the issues of training a semi-supervised VAE is that the marginalization over $\qphi(y|\x)$ in eq.~\eqref{eq:unlabeled_obj} can be computationally expensive. This marginalization can be avoided by using Gumbel-Softmax \cite{Gumbel,gumbel2}, a continuous distribution on the probability simplex that approximates a categorical sample and can be smoothly annealed (through a temperature parameter) to the categorical distribution. Gumbel-Softmax is reparameterizable so that the gradient of the loss function can be propagated back through the sampling step $y \sim \qphi(y|\x)$ for single-sample gradient estimation.

\subsubsection{Regularization}
The lower bound for labeled data can be rewritten as
\begin{align}
    &\mathcal{L}_{\thetab, \phib}(\x,y) = \ex_{\qphi(\z|\x,y)}\left [\log \frac{\ptheta(\x,y,\z)}{\qphi(\z|\x,y)} \right ]\nonumber\\
    & \qquad = \ex_{\qphi(\z|\x,y)}\Big[ \log \ptheta(\x  | \z, y) \Big] + \log p(y) - D_{KL}(\qphi(\z|\x,y)||p(\z))\nonumber
\end{align}
where $\log p(y)$ is a constant, the first term can be interpreted as expected negative reconstruction error, and the last term is the negative KL divergence from the prior to the approximate posterior. Similarly, we can express the bound for unlabeled data as follows:
\begin{align*}
    &\mathcal{U}_{\thetab,\phib}(\x) = \ex_{\qphi(\z, y|\x)}\Big[ \log \ptheta(\x  | \z, y) \Big] - D_{KL}(\qphi(\z,y|\x)||p(\z, y))
\end{align*}
In both cases, the KL divergence acts as a regularization term that encourages the approximate posterior to be close to the prior, thereby constraining the amount of information encoded in the latent variables. The overall lower bound~\eqref{eq:comb_objective} thus trades off reconstruction error with this regularization term.
%
When training a VAE, we can control such trade-off in order to favor more accurate reconstructions or more constrained latent space, by simply multiplying the KL term by a factor $\beta>0$ as proposed in~\cite{higgins2017beta}. 
Similarly, we found it beneficial in practice to scale the entropy of $q_{\phi}(y | \mathbf{x})$ in eq.~\eqref{eq:unlabeled_obj} by a factor $\gamma > 1$. Intuitively, the entropy term acts as a regularizer in the classifier by encouraging $q_{\phi}(y | \mathbf{x})$ to have high entropy: the amplification of this term helps to further reduce overfitting in the classifier.



\section{Data and models}
The BraTS 2019 challenge is composed of a training, a validation and a test set.
The training set 
is composed of 335 delineated tumor images, in which 210 images have survival labels. The validation set is composed of 125 non-delineated images without survival labels,
in which only 29 images with resection status of GTR (i.e., Gross Total Resection) are part of the online evaluation platform (CBICA's Image Processing Portal).
Finally, the test set will be made available to the challenge participants during a limited time window, and the results will be part of the BraTS 2019 workshop. 


In all our experiments we performed 3-fold cross-validation by randomly
splitting the BraTS 2019 training set with survival labels into a ``private'' training (75\%) and validation set (25\%)
in each fold,
in order to have an alternative to the online evaluation platform.
This help us having a 
more informative
indication of the model performance, since the online evaluation platform includes just 29 cases (vs.~$53$ 
cases in our private validation sets). 
With this set-up, which we call {\bf S0} in the remainder, we effectively trained the model on a training set of $\mathbf{N_l=157}$ and $\mathbf{N_u=125}$ 
for each of the three cross-validation folds.
These models were subsequently tested on their corresponding private validation sets of 53 subjects, as well as on the standard BraTS 2019 validation set of 29 subjects.

In order to evaluate just how much the proposed method is able to learn from \emph{unlabeled} data (i.e., subjects with tumor delineations but no survival time information),
we used three open-source methods \cite{seg1,seg2,seg3} to automatically segment both the entire 
BraTS 2019
training and validation sets in order to have many more unlabeled training subjects available. 
We further augmented these unlabeled data sets by flipping the images in the coronal plane. 
With this new set-up, which we call {\bf S1}, we then trained the model on an ``augmented'' private training set of 
$\mathbf{N_l=157}$ and $\mathbf{N_u=2268}$ for each of the three cross-validation folds.
Ideally, 
dramatically
increasing the set of unlabeled data points this way should help the model learn to better encode tumor representations,
thereby increasing classification accuracy.

\section{Implementation}
\label{sec:implementation}

We implemented the encoder $\qphi(\z|\x, y)$, the decoder $\ptheta({\x} | \z, y)$ and the classifier $\qphi(y|\x)$ all as deep convolutional networks using PyTorch \cite{pytorch}. The segmentation volumes provided in the BraTS challenge have size $240\!\times\!240\!\times\!155$, but since large parts of the volume are always zero, we cropped the volumes to $146\!\times\!188\!\times\!128$ without losing any tumor voxels. We further reduced the volume by a factor of 2 in all dimensions, resulting in a shape of $73\!\times\!94\!\times\!64$, roughly a 95\% overall reduction in input image size. This leads to faster training and larger batches fitting in memory, while losing minimal information.

We optimized the model end-to-end with Adam optimizer~\cite{adam}, using a batch size of 32, learning rate $2\cdot 10^{-5}$, latent space size 32, $\alpha = 10^{-5} \cdot D \approx 4.4$ with $D$ the data dimensionality (number of voxels), $\beta$ from $0$ to $6 \cdot 10^{3}$ in $3 \cdot 10^{4} $ steps, $\gamma = 50$, and exponentially annealing the Gumbel-Softmax sampling temperature from $1.0$ to $0.2$ in $5 \cdot 10^{4}$ steps. Hyperparameters were found by grid search, although not fine-tuned because of the computational cost.
The total number of parameters in the model is around $2.7 \times 10^6 $.


\subsection{Network architecture}
\label{ssec:architecture}

\begin{figure}[t!]
    \centering
    \includegraphics[width=.95\textwidth]{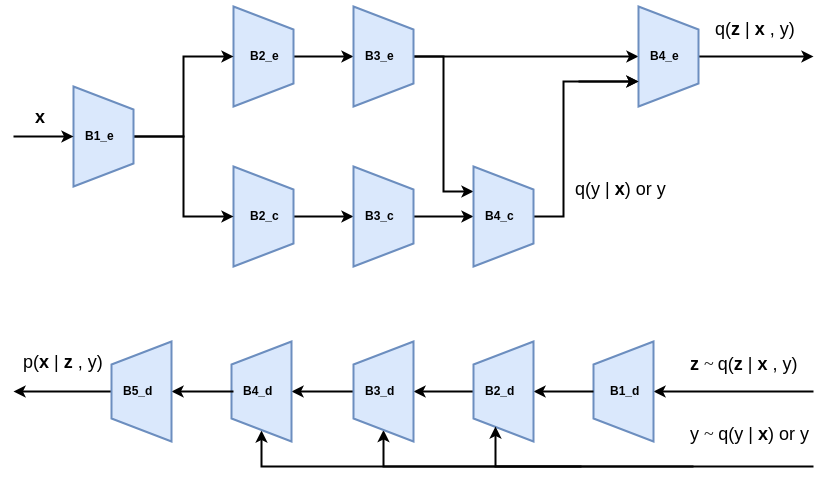}
    \caption{Networks architectures: encoder, decoder and classifier architectures.}
    \label{fig:semiVAENet}
\end{figure}

The three networks consist of 3D convolutional layers, with the exception of a few fully connected layers in the classifier. There are nonlinearities (Scaled Exponential Linear Units,~\cite{selu}) and dropout~\cite{dropout} after each layer, except when noted. What follows is a high-level description of the network architecture, represented in diagrams in Figure~\ref{fig:semiVAENet}
. For more details, the code is available at \url{https://github.com/sveinnpalsson/semivaebrats}.

The inference network consists of a convolutional layer (B1\_e) with large kernel size and stride (7 and 4, respectively), followed by two residual blocks~\cite{residual} (B2\_e and B3\_e). The input to each block is processed in parallel in two branches, one consisting of two convolutional layers, the other of average pooling followed by a linear transformation (without nonlinearities). The results of the two branches are added together. The output of the first layer is also fed into the classifier network, which outputs the class scores (these will be used to compute the classification loss for labeled data). A categorical sample from $q_{\phi}(y | \mathbf{x})$ is drawn using the Gumbel-Softmax reparameterization given the class scores, and is embedded by a fully connected layer into a real vector space. Such embedding is then concatenated to the output of the two encoder blocks, so that the means and variances
of the approximate posterior $q_{\phi}(\mathbf{z} | \mathbf{x}, y)$, that are computed by a final convolutional layer, are conditioned on the sampled label.
The classifier consists of two residual blocks similar to the ones in the encoder (B2\_c and B3\_c), followed by two fully connected layers (B4\_c).

The decoder network consists of two convolutional layers (B1\_d and B2\_d), two residual blocks similar to those in the encoder (B3\_d and B4\_d), and a final convolution followed by a sigmoid nonlinearity (B5\_d). In the decoder, most convolutions are replaced by transposed convolutions (for upsampling), and pooling in the residual connections is replaced by nearest neighbour interpolation. The input to the decoder network is a latent vector $\mathbf{z}$ sampled from the approximate posterior. The embedding of $y$, computed as in the final stage of the inference network, is also concatenated to the input of each layer (except the ones in the middle of a block) to express the conditioning of the likelihood function on the label. Here, the label is either the ground truth (for labeled examples) or a sample from the inferred posterior (for unlabeled examples).

\section{Results}

\subsection{Conditional generation}

\begin{figure}
    \centering
    \includegraphics[width=.95\textwidth]{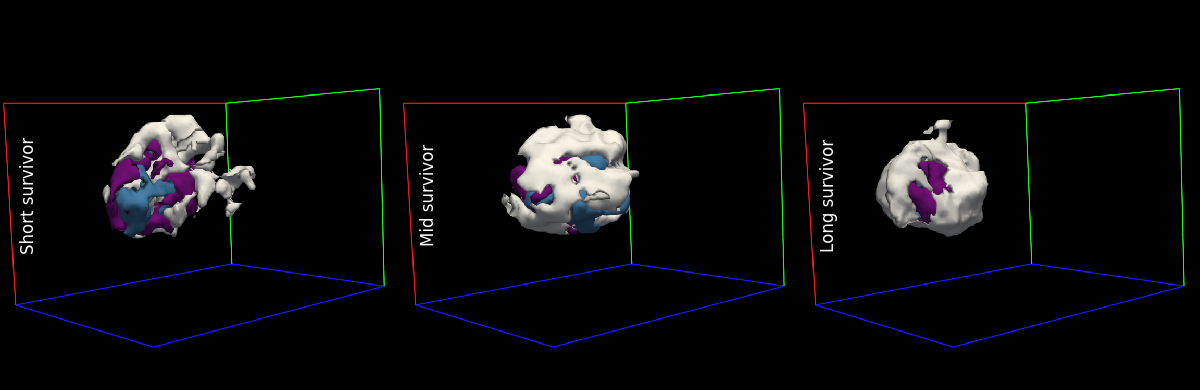}
    \caption{Generated tumor from $\ptheta(\x | y, \z)$ where we sampled ${\bf z}$ from $\mathcal{N}({\bf z} | {\bf 0}, {\bf I})$ and we varied $y$ between short survivor, mid survivor and long survivor.}
    \label{fig:genTumor}
\end{figure}

We visually tested whether the decoder $\ptheta({\x} | y, \z)$ is able to generate tumor-like images after training, and whether it can disentangle the classes. For this purpose we sampled ${\z}$ from $\mathcal{N}({\z} | {\bf 0}, {\bf I})$ and varied $y$ between the three classes, namely, short survivor, mid survivor and long survivor. Figure~\ref{fig:genTumor} shows the three shapes generated accordingly by one of the models trained in set-up {\bf S0}.
From the images we can see that the generated tumor for the short survivor class has an irregular shape with jagged edges while the long survivor generated tumor has a more compact shape with rounded edges.

\subsection{Quantitative evaluation}

All the classification accuracies are reported with binomial confidence interval with normal approximation \cite{CI}, defined as
\begin{align*}
    \large\displaystyle a \pm {{z^{*}}}\sqrt{\frac{a(1-a)}{n}}
\end{align*}
where $a$ is the classification accuracy, $z^{*}=1.96$ is the critical value with confidence level at 95\% and $n$ is the number of subjects.
In Table~\ref{tab:3fold} we show the classification accuracy of the proposed method on the ``private'' validation set of $53$ subjects for each of the three cross-validation folds, both for the set-up with fewer ({\bf S0}) and more ({\bf S1}) unlabeled training subjects.
The corresponding results based on 
the online evaluation platform (29 validation subjects)
are summarized in Table~\ref{tab:BraTSVal}, where we submitted the majority vote for survival group prediction across the three models trained in the cross-validation folds. The online evaluation platform takes the estimated number of days as input and returns the accuracy along with mean- and median squared error and Spearman's rank correlation coefficient. To make these predictions we input the average survival from each class. Our scores on the challenge leaderboard for set-up {\bf S0} are as follows: 37.9\% accuracy, 111214.828 mean squared error, 51076.0 median squared error and a correlation of 0.36. When testing the models we found that they are insensitive to the segmentation method used to produce the input.


\begin{table}
\centering
 \caption{Classification accuracies [\%] for both set-ups on the ``private'' validation set for each of the three cross-validation folds.}
 \label{tab:3fold}
\tabcolsep=0.185cm
 \begin{tabular}{l | c | c | c || c } 
 Set-up & Fold 1 & Fold 2 & Fold 3 & Avg \\ [0.5ex] 
 \hline
 {\bf S0}  & $42.18 \pm 13.30$ & $35.90 \pm 12.91$ & $39.53 \pm 13.16$ & $39.20 \pm 7.59$\\
 {\bf S1}  & $47.55 \pm 13.45$ & $41.13 \pm 13.40$ & $42.91 \pm 13.32$ & $43.86 \pm 7.71$\\
 \end{tabular}
\end{table}

\begin{table}
\centering
 \caption{Classification accuracies [\%] for both set-ups on the BraTS 2019 online evaluation platform.}
 \label{tab:BraTSVal}
\tabcolsep=0.185cm
 \begin{tabular}{l | c } 
 Set-up & Majority voting \\ [0.5ex] 
 \hline
 {\bf S0}  &  $37.90 \pm 17.57$\\
 {\bf S1}  &  $31.00 \pm 16.83$\\
 \end{tabular}
\end{table}

The results show that in none of the experiments our model achieved a significant improvement over always predicting the largest class, which constitutes around 40\% of the labeled cases.

\section{Discussion and conclusions}

In this paper we evaluated the potential of a semi-supervised deep generative model for classifying brain tumor patients into three overall survival groups, based only on tumor segmentation masks.
The main potential advantages of this approach are (1)
its in-built invariance to MR intensity variations when different scanners and protocols are used,
enabling wide applicability across clinics;
and (2)
its ability to learn from 
unlabeled data, which is much more widely available than fully-labeled data.
%

We compared two different set-ups: one where fewer unlabeled subjects were available for training,
and one where 
their number was (largely artificially) increased
using automatic segmentation and data augmentation.
Although the latter set-up increased classification performance in our ``private'' experiments, this increase did not reach statistically significant levels 
and was not replicated on the small BraTS 2019 validation set.
We demonstrated visually that the proposed model effectively learned class-specific information, but overall failed to achieve classification accuracies significantly higher than predicting always the largest class.

The results described here are only part of a preliminary analysis. More real unlabeled data, obtained from truly different subjects pooled across treatment centers,
and more clinical covariates of the patients, such as age and resection status, 
may be necessary to reach better classification accuracies. Future work may also involve stacking hierarchical generative models to further increase the classification performance of the model~\cite{VAESEMIKingma}. 

\section{Acknowledgements}
This project was funded by the European Union’s Horizon 2020 research and innovation program under the Marie Sklodowska-Curie project TRABIT (agreement No 765148).

%
%
%
\bibliographystyle{unsrt}
\bibliography{refs}

\end{document}